# Active Model Selection

**Omid Madani**
Yahoo! Research Labs
74 N. Pasadena Ave,
Pasadena, CA 91101
omid.madani@overture.com

**Daniel J. Lizotte** and **Russell Greiner**
Dept. of Computing Science
University of Alberta
Edmonton, T6J 2E8
{dlizotte | greiner}@cs.ualberta.ca

## Abstract

Classical learning assumes the learner is given a labeled data sample, from which it learns a model. The field of Active Learning deals with the situation where the learner begins not with a training sample, but instead with resources that it can use to obtain information to help identify the optimal model. To better understand this task, this paper presents and analyses the simplified "(budgeted) active model selection" version, which captures the pure exploration aspect of many active learning problems in a clean and simple problem formulation. Here the learner can use a fixed budget of "model probes" (where each probe evaluates the specified model on a random indistinguishable instance) to identify which of a given set of possible models has the highest expected accuracy. Our goal is a policy that sequentially determines which model to probe next, based on the information observed so far. We present a formal description of this task, and show that it is NP-hard in general. We then investigate a number of algorithms for this task, including several existing ones (eg, "Round-Robin", "Interval Estimation", "Gittins") as well as some novel ones (*e.g.*, "Biased-Robin"), describing first their approximation properties and then their empirical performance on various problem instances. We observe empirically that the simple biased-robin algorithm significantly outperforms the other algorithms in the case of identical costs and priors.

## 1 Introduction

Learning tasks typically begin with a data sample — *e.g.*, symptoms and test results for a set of patients, together with their clinical outcomes. By contrast, many real-world studies begin with *no* actual data, but instead with an idea and a budget — funds that can be used to collect the relevant information. For example, one study has allocated $2 million to develop a system to diagnose cancer subtypes, based on a battery of tests on collected tissue, each test with its own (known) costs and (unknown) discriminative powers [Pol]. Given our goal of identifying the most accurate model, what is the best way to spend the $2 million? Should we indiscriminately run every test on every tissue, until exhausting the budget? ... or selectively, and dynamically, determining which tests to run on which tissue? We call this general problem *budgeted learning*.

In that study, the eventual model will be allowed to perform tests to identify cancer types. To better understand the fundamentals of this general problem, we investigate a simpler "(budgeted) active model selection" variant, where the tissues are indistinguishable, and the goal is to identify which single test to apply to *all* (future) tissues. That is, we have a fixed set of possible diagnostic tests — "models" — and the learner's task is to select exactly one of them. As above, the learner can work "actively", sequentially deciding (at learning time) which test should be applied to which tissue, to help identify which test is better is general.

To simplify our notation, we will view this problem as the "coins problems": We are given $n$ (distinguishable) coins with unknown head probabilities. We are allowed to sequentially specify a coin to flip, then observe the outcome of this flip, but only for a known, fixed number of flips. After this trial period, we have to declare a winner coin. Our goal is to pick the coin with the highest head probability from among the coins. However, considering the limits on our trial period, we seek a strategy for coin flipping that, on average, leads to picking a coin that is as close to the best coin as possible.

There is a tight relation between active model selection (*i.e.*, identifying the best coin) and identifying the most discriminative test or feature: the head probability of a coin is a measure of quality, and corresponds to the discrimination power (*e.g.*, accuracy) in the feature selection problem. The "features" may actually be more sophisticated classifiers such as decision trees, with known expected costs but unknown accuracies. In the latter case however, we are ignoring the fact that the different models may share features and hence be correlated. This active model selection problem is an abstraction applicable to other scenarios, such as determining the best parameter settings for a program given a deadline that only allows a fixed number of runs; or choosing a life partner in the bachelor/bachelorette TV show where time is limited. Finally, note that the hardness results and the algorithmic issues that we identify in this work also apply to the more general budgeted classifier learning problems [LMG03].

The first challenge in defining the budgeted active model selection problem is to formulate the objective, to obtain a well-defined and satisfactory notion of *optimality* for the complete range of budgets. We do this by assigning priors over coin quality, and by defining a measure of regret for choosing a coin as a winner. We describe *strategies* (for determining which coin to flip in each situation), and extend the definition of regret to strategies. The computational task is then reduced to identifying a strategy with minimum regret, among all strategies that respect the budget.

We address the computational complexity of the problem, showing that it is in PSPACE, but also NP-hard under different coin costs. We establish a few properties of optimal strategies, and also explore where some of the difficulties may lie in computing optimal strategies, *e.g.*, the need for contingency in the strategy, even when all coins have the same cost. We investigate the performance of a number of algorithms empirically and theoretically, by defining and motivating constant-ratio approximability. The algorithms include Interval Estimation and (adapted) Gittin indices [Kae93, BF85], obvious ones such as Round-Robin, as well as novel ones that we propose based on our knowledge of problem structure. One such algorithm, "Biased-Robin", works especially well for the case of identical costs and priors. The paper also raises a number of intriguing open problems.

The main contributions of this paper are:

**1.** Precisely defining the basic active model selection problem in this space, as a problem of sequential decision making under uncertainty.

**2.** Addressing the computational complexity of the problem, highlighting important issues both for optimality and approximability. Empirically comparing a number of obvious, and not so obvious, algorithms, towards determining which work most effectively.

**3.** Providing, in closed-form, the expected regret, under uniform priors, of an obvious algorithm: Round-Robin (and variants).

Section 2 defines the coins problem and presents its computational complexity. Section 3 defines the constant-ratio approximation property, describes the algorithms we study — both familiar and novel — and explores approximability. Section 4 empirically investigates the performance of the algorithms over a range of inputs and Section 5 discusses related work, distinguishing this work differs from related notions, such as bandit problems, experimental design and on-line learning. For the proofs and derivations, extended explanations, additional empirical results, and *animated* algorithms, please see [MLG04].

## 2 The Coins Problem

We are given:

• A collection of $n$ independent coins, indexed by the set $\mathcal{I}$, where each coin is specified by a query (flip) cost and a probability density function (prior) over its head probability. The priors of the different coins are independent, and they can be different for different coins.

• A budget $b$ on the total allowed cost of querying.

We assume the tail and the head outcomes will correspond to receiving no reward and a fixed reward (1 unit) respectively, at performance tain. We are allowed a trial/learning period, constrained by the budget, for the sole purpose of experimenting with the coins, *i.e.*, we do not collect rewards in this period. At the end of the period, we are allowed to pick only a single coin for all our future flips (reward collection).

Let the random variable $\Theta_i$ denote the head probability of a coin $c_i$, and let $w_i(\Theta_i)$ be the density over $\Theta_i$. Note that the densities can change based on the results of the coin flips. We first address the question of which coin to pick at the end of the learning period, *i.e.*, when the remaining budget allows no more flips. The expected head probability of coin $c_i$, aka the *mean* of coin $c_i$, is: $E(\Theta_i) = \int_0^1 \theta\, w_i(\theta)\, d\theta$. The coin to pick is the one with the highest mean $\mu_{max} = \max_{i \in \mathcal{I}} E(\Theta_i)$, which we denote by $i^*$. The motivation for picking coin $i^*$ is that flipping such a coin gives an expected reward no less than the expected reward obtained from flipping any other coin.

We can now define the measure of error that we aim to minimize. Let $i_{max}$ be a coin with the highest head probability, and let $\Theta_{i_{max}} = \Theta_{max} = \max_{i \in \mathcal{I}} \Theta_i$, be the random variable corresponding to the head probability of $i_{max}$. For example, in case of two coins with $\Theta_1 = 0.1$ and $\Theta_2 = 0.4$, $\Theta_{max} = 0.4$. In general, as these $\Theta_i$ are random variables, $\Theta_{max}$ is also a random variable, with expectation $E(\Theta_{max}) = \int_{\vec{\theta}} (\max_{i \in \mathcal{I}} \theta_i) \prod_i w_i(\theta_i)\, d\vec{\theta}$. For example, given $n$ coins drawn with uniform priors, we observe $E(\Theta_{max}) = \frac{n}{n+1}$ [MLG04]. The (expected) *regret* from picking coin $c_i$ is then $E(r(i)) = E(\Theta_{max} - \Theta_i) = E(\Theta_{max}) - E(\Theta_i)$, *i.e.*, the average amount by which we would have done better[1] had we chosen coin $i_{max}$

[1] In the standard bandit framework [BF85] where exploitation is also important, the term "regret" is commonly used to refer to a conceptually similar but slightly different quantity. We use the

instead of coin $c_i$. Observe that we minimize regret by picking coin $i^*$. Thus the (expected) minimum regret is $E(\Theta_{max}) - \mu_{max}$. Note that the (minimum) regret has an interesting easy to remember form: it is the difference between two quantities that differ in the order of taking maximum and expectation:

$$E(\Theta_{max}) = E(\max_{i\in\mathcal{I}} \Theta_i)$$
$$\mu_{max} = \max_{i\in\mathcal{I}} E(\Theta_i)$$

### 2.1 Strategies

Informally, a strategy is a prescription of which coin to query at a given time point. In general, such a prescription depends on the objective (minimizing regret in our case), the outcomes of the previous flips or equivalently the current posterior densities over head probabilities (*i.e.*, the current *belief state*), and the remaining budget. Note that after a flip of coin $c_i$, and observing outcome $o$, the density over its head probability is updated using Bayes formula, and the probability of a head outcome for flipping coin $i$ becomes $E(\Theta_i|o)$. Equivalently, we simply update the density for coin $i$ according to the observation, and the expectation is taken over the current density (see Section 2.3).

A strategy may be viewed as a finite, rooted, directed tree where each leaf node is a special "stop" node, and each internal node corresponds to flipping a particular coin, whose two children are also strategy trees, one for each outcome of the flip (see Figure 1). We will only consider strategies respecting the budget, *i.e.*, the total cost of coin flips along any branch may not exceed the budget. Thus the set $S$ of strategies to consider is finite though huge: $n^{2^b-1}$ assuming unit costs. Associated with each leaf node $j$ of a strategy is the regret $r_j$, computed using the belief state at that node, and the probability of reaching that leaf $p_j$, where $p_j$ is the product of the transition probabilities along the path from root to that leaf. We therefore define the *regret* of a strategy to be the expected regret over the different outcomes, or equivalently the expectation of regret conditioned on execution of $s$, or $E(r|s)$:

$$Regret(s) = E(r|s) = \sum_{j\in\text{Tree Leafs}} p_j r_j$$

An *optimal strategy* $s^*$ is then one with minimum regret: $s^* = \arg\min_{s\in S} Regret(s)$. Figure 1 shows an optimal strategy[2] for the case of $n \geq 4$ coins with uniform priors, and a budget of $b = 3$. We have observed that optimal strategies for identical priors typically enjoy a similar pattern (with some exceptions): their top branch (*i.e.*, as long as the outcomes are all heads) consists of flipping the same coin, and the bottom branch (*i.e.*, as long as the outcomes are all tails) consists of flipping the coins in a Round-Robin fashion; see Biased-Robin (Section 3.1.3 below).

same term as it best describes the objective in this setting as well.

[2]Note there always exists an optimal strategy that is deterministic. A way to see this is to realize that the coins problem is a special finite-horizon fully observable MDP (see Section 5).

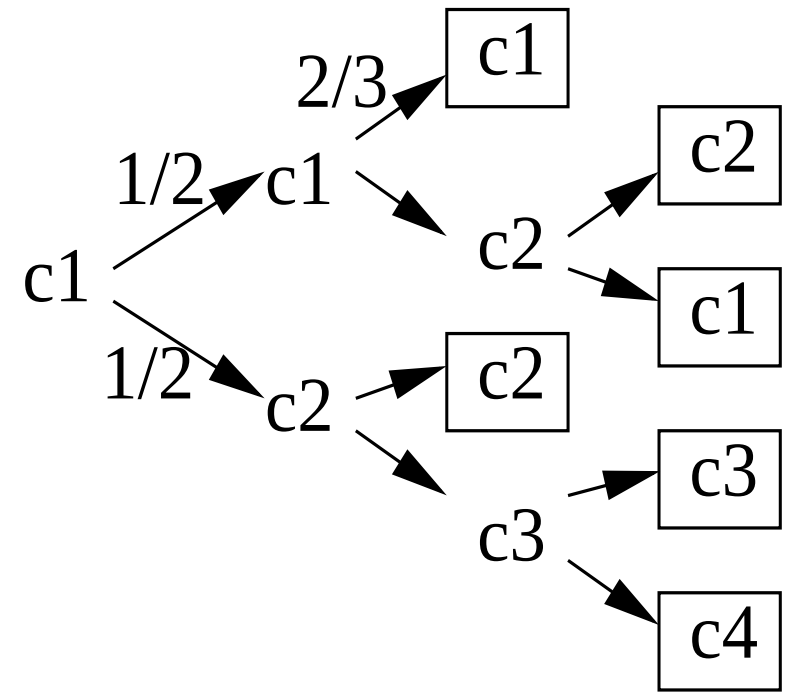

Figure 1: The optimal strategy tree for budget $b = 3$ on identical uniform priors, with 4 (or more) coins. Top branches correspond to head outcomes. Some branches terminate early as the coin to choose (boxed) is already determined.

### 2.2 The Computational Problem

Our overall goal is to execute flips according to some *optimal strategy* $s^*$. Three relevant computational problems are outputting (1) an optimal strategy, (2) the best coin to flip now (the first action of the optimal strategy), or (3) the minimum regret. As the optimal strategy may be exponential in the input size, when we talk about the coins problem in a formal sense (*e.g.*, Theorem 1), we mean the problem of computing the first action of an optimal strategy.

### 2.3 Beta Densities

In our experiments, we will be using the family of Beta densities to model the density over a coin's head probability $\Theta$. This family is particularly convenient for representing and updating priors [Dev95]. Briefly, a Beta density is a two-parameter function, $B(\alpha_1, \alpha_2)$ of the form $C\Theta^{\alpha_1-1}(1-\Theta)^{\alpha_2-1}$, where $C$ is a normalizing constant and, in this paper, $\alpha_1$ and $\alpha_2$ are positive integers. $B(1, 1)$ correponds to the uniform prior. The density $\Theta \sim B(\alpha_1, \alpha_2)$ has mean $E(\Theta) = \mu = \alpha_1/(\alpha_1+\alpha_2)$, and standard deviation $STD(\Theta) = \sqrt{\frac{(1-\ )}{\alpha_1+\alpha_2+1}}$. After a heads (resp. tails) outcome, $B(\alpha_1, \alpha_2)$ is updated to $B(\alpha_1+1, \alpha_2)$ (resp. $B(\alpha_1, \alpha_2+1)$). See [MLG04] for more details and examples.

### 2.4 Computational Complexity

The coins problem is a problem of sequential decision making under uncertainty, and similar to many other such problems [Pap85], one can verify that it is in PSPACE, as long as we make the assumption that the budget is bounded by a polynomial function of the number of coins and density updates and regret computations given the current densities of the coins can be computed in PSPACE. Moreover, if we assume the number of coins $n$ is constant, then the obvious dynamic program can solve this problem in time polynomial in the budget (if exponential in $n$). However, in general, the problem is NP-hard:

**Theorem 1** *The coins problem is in PSPACE and NP-hard [MLG04].*

The proof reduces the Knapsack Problem to a special coins problem where the coins have different costs, and discrete priors with non-zero probability at head probabilities $0$ and $1$ only. It shows that maximizing the profit in the Knapsack instance is equivalent to maximizing the probability of finding a perfect coin, which is shown equivalent to minimizing the regret. The reduction reveals the packing aspect of the budgeted problem. It remains open whether the problem is NP-hard when the coins have unit costs and/or uni-modal distributions. The next section discusses some difficulties in computing optimal strategies even in this restricted case, and explores the related issue of approximability.

## 3 Problem Structure and Algorithm Design

The following are a few simplifying and somewhat intuitive properties.

**Proposition 2**
**1.** $E(\Theta_{\max}|s) = E(\Theta_{\max})$*, therefore, the regret of a strategy* $s$ *is,*

$$regret(s) \quad = \quad E(\Theta_{\max}) \ - \ E(\mu_{max}|s),$$

*where* $E(\Theta_{\max}|s)$ *and* $E(\mu_{max}|s)$ *denote respectively the conditional expectations of max random variable* $\Theta_{\max}$ *and maximum mean* $\mu_{max}$*, conditioned on the execution of strategy* $s$ *(i.e., the expectations over all the outcomes of executing strategy* $s$*).*
**2.** *More information cannot hurt: For any strategy* $s$, $E(\mu_{max}|s) \geq E(\mu_{max})$*, therefore regret from using any strategy* $s$ *is not greater than the current regret.*
**3.** *No need to query useless coins: Assume that under any outcome (i.e., the execution of any strategy respecting the budget), there is some coin whose mean is at least as high as coin* $c_i$*. Then there exists an optimal strategy tree that never queries coin* $c_i$*.* ∎

The first property follows from the fact that an expectation over an expectation does not change the value. Thus, minimizing regret is equivalent to maximizing the expected highest mean $E(\mu_{max}|s)$, and the latter is often easier to compute (*e.g.*, see Section 3.1.2). The second and third properties are established by induction on strategy tree height, and considering a few special cases. We conclude that the optimization problem boils down to computing a strategy that maximizes the expectation over the highest mean; $s^* = \text{argmax}_{s \in S}\{E(\mu_{max}|s)\}$.

It follows that in selecting the coin to flip, two significant properties of a coin are the magnitude of its current mean, and the spread of its density (think "variance"), that is how changeable its density is if it is queried: if a coin's mean is too low, it can be ignored by the above result, and if its density is too peaked (imagine no uncertainty), then flipping it may yield little or no information (the expectation $E(\mu_{max}|s)$ may not be significantly higher than $E(\mu_{max})$). However, the following simple, two coin example shows that the optimal action can be to flip the coin with the lower mean and lower spread!

**Example 1:** Assume coin $c_1$ has $B(1,2)$ prior, and coin $c_2$ has $B(1,3)$; thus $c_1$ has a higher mean and a lower spread than $c_2$. But the optimal strategy for a budget of one starts by flipping $c_2$. (To see why, note that flipping $c_1$ does not change our decision under either of its two outcomes — $c_1$ will be the winner — and thus the $E(\mu_{max})$ equals the current highest mean value of $1/3$, while flipping $c_2$ affects the decision, and the expectation of the $\mu_{max}$ given that $c_2$ is queried is slightly higher — $1/4 \times 2/5 + 3/4 \times 1/3$.) ∎

It would be nice if there was some "local" property of a coin — ideally a single scalar value — that was sufficient to identify which coin to flip, at each time step. Unfortunately...

**Example 2: Context Sensitivity:** Suppose you have a budget of $1$ to decide between the two coins, $c_1$ with $\Theta_1 \sim B(1,1)$ and $c_2$ with $\Theta_2 \sim B(5,3)$, Here, it is clear that you should flip $c_1$. *If* this decision was based on a single number associated with $c_1$ (resp., $c_2$), then adding a third coin $c_3$ could not change the order between $c_1$ and $c_2$; and in particular, a final budget of $1$ flip would ever be given to $c_2$. However, this will happen if $c_3$ is distributed with $\Theta_3 \sim B(17,9)$! ∎

The next example shows that the optimal strategy can be *contingent* — *i.e.*, the optimal flip at a given stage depends on the outcomes of the previous flips.

**Example 3: Contingent:** With the three unit-cost coins $c_1 \sim B(1,1)$, $c_2 \sim B(5,2)$ and $c_3 \sim B(21,11)$, and a budget of $b = 2$, the optimal strategy is to flip coin 1, and if the outcome is heads, flip it again. If the outcome is tails, flip coin 2. It can be verified that the best strategies when starting with flipping $c_1$, or $c_2$, or $c_3$, give expected highest means of $0.731$, $0.725$, and $0.723$, respectively. ∎

Note that the examples involve identical costs. These observations suggest that optimization may remain hard even in the identical costs case. However, the difference between the optimal regret and regret of a simple algorithm, which for example ignores contingency, may not be significant. We explore approximability and candidate approximation algorithms in the next subsection.

### 3.1 Algorithms and Approximability

Consider an algorithm $\mathcal{A}$ that given the input, outputs the next action to execute. We call algorithm $\mathcal{A}$ a *(constant-ratio) approximation algorithm* if there is a constant $\ell \in [1,\infty)$ (independent of problem size), such that given any problem instance, if $r^*$ is the optimal regret for the prob-lem, the regret $r(\mathcal{A})$ from executing actions prescribed by $\mathcal{A}$ is bounded by $\ell \times r^*$. A constant-ratio approximation is especially desirable, as the quality of the approximation does not degrade with problem size. Of course we seek an approximation algorithm (preferably with low $\ell$) that is also efficient (polynomial time in input size). We next describe a number of plausible algorithms, and explore whether or not they are approximation algorithms. For an animation of several of these algorithms, please see [MLG04].

#### 3.1.1 Round-Robin, Random, and Greedy Algorithms

The *Round-Robin* algorithm simply flips the coins in a Round-Robin fashion, ***i.e.***, flips coin $i = (t-1 \bmod n)+1$, at time $t = 1, 2, \cdots$. The *Random* algorithm at each decision point simply picks a coin uniformly at random and flips it. These algorithms are plausible algorithms, at least initially in the trial period, and they are a standard protocol in clinical trials (***e.g.***, [Pol]). The third algorithm we consider is the *Constant-budget* algorithm: For a small constant $k$ (independent of $n$ and $b$), it computes the optimal strategy for that *smaller* budget $k$, and flips the first coin of such a strategy. (Given the outcome, it then computes the optimal strategy from this new state, with the decremented budget, etc.) We shall refer to the algorithm as simply *Greedy* when $k = 1$. Perhaps it is not hard to see these algorithms are suboptimal, but we can say more:

**Proposition 3 ([MLG04])** *For algorithm $\mathcal{A} \in \{$ Round-Robin, Random, Constant-Budget $\}$, for any constant $\ell$, there is a problem with minimum regret $r^*$, on which $r(\mathcal{A}) > \ell \times r^*$.* ∎

#### 3.1.2 Allocational Algorithms (including SCLA)

An allocational strategy is specified by the number of flips assigned to each coin. For example, given a budget of $5$, an allocation may specify that coin 1 should be flipped twice, coin 2 flipped once, and coin 3 twice (and all other coins 0 times). Notice this allocation does not specify when to flip a coin (any coin with positive allocation may be flipped first), and it is not contingent. The attraction of allocational strategies is that they are compactly represented. We also show that they are *efficiently* evaluated: the expected highest mean of an allocational strategy can be computed in time polynomial in $nb$ [MLG04]. With an equal allocation of $a$ flips to every coin (***e.g.***, Round-Robin when $b = a\,n$) and under uniform priors, the expression for the regret (***i.e.***, $E(\Theta_{\max}) - E(\mu_{max}|s)$) further simplifies to:

$$\frac{n}{n+1} - \sum_{h=0}^{a} \frac{(h+1)^n - h^n}{(a+1)^n} \frac{h+1}{a+2}. \quad (1)$$

In addition to Round-Robin, we also consider an extreme restricted version of a dynamic allocational strategy, the single-coin allocational strategy, aka (single-coin) *look-ahead* algorithm (SCLA): at each time point, for each coin $c_i$, the look-ahead algorithm considers allocating *all* of the remaining flips to coin $c_i$, computes the expected highest mean from each single-coin allocation, and flips a coin that gives the largest expected highest mean. (That is, given the initial budget of $b$, it computes the coin to flip; $i_b^{SCLA} = SCLA(b)$. Given the outcome of this flip, it then computes, and flips, the coin $i_{b-1}^{SCLA} = SCLA(b-1)$, and so forth.) This computation can be done in polynomial time — $O(n\,b)$ at every time point. While this algorithm may not be an approximation algorithm, with specially designed non-identical priors [MLG04], we will see empirically that it performs fairly well.

#### 3.1.3 Biased-Robin

The Biased-Robin algorithm is similar to Round-Robin, except that it keeps flipping the same coin as long as it gives heads. Thus, Biased-Robin begins by choosing coin $c_1$ and flipping it. It keeps flipping the currently chosen coin until the coin gives a tail outcome, in which case it chooses the next coin, wrapping around and starting with coin $c_1$ whenever a flip of coin $c_n$ gives a tail outcome. The Biased-Robin algorithm is inspired by the overall pattern that we observed in the optimal strategy tree for the case of identical priors (Figure 1). It is also a generalization of Robbins' "play the winner" strategy for two Bernoulli arms [Rob52, Zel69] (if a success occurs on one arm, the arm is repeated, while if a failure occurs, then the other arm is tried), and the same strategy arises when we are searching for a perfect model (see ***e.g.***, [SG95]). However, Biased-Robin is not optimal, as the optimal strategy does not follow this pattern completely. Exceptions occur, for example, when the remaining budget is low [MLG04]. Like Round-Robin, Biased-Robin does not take either the priors nor the budget into account, and it is any-time. But, somewhat unexpectedly, we observe empirically that it does very well.

#### 3.1.4 Interval Estimation

The Interval Estimation algorithm [Kae93] attempts to account for the uncertainty over the performance of a model (coin) by flipping the coin that has the highest "reasonably likely" performance, where reasonably likely performance is defined as the top of the 95% confidence interval, which is the sum of the current mean of the coin and a multiple $\gamma = 1.96$ of the standard deviation of the distribution $STD(\Theta)$ over the coin's head probability. At each time point flip a coin $i^{IE}$ with the highest such confidence interval,

$$i^{IE} \quad = \quad \arg\max_{i \in \mathcal{I}} \{E(\Theta_i) + \gamma \times STD(\Theta_i)\}$$

(Note the higher the tolerance $\gamma$, the more the algorithm is biased towards coins with high spread over their performance, while for $\gamma = 0$, the algorithm flips the coin with current highest mean — ***i.e.***, reduces to pure exploitation.)

#### 3.1.5 Gittins Indices

Our "active model selection" problem is obviously related to the standard, well-studied "Bandit Problem" [BF85]: you are facing a set of $n$ "one-armed bandits" (aka "slot machines"), each with some fixed but unknown expected payoff. At each time, you decide which arm to pull, then receive a payoff drawn from the output distribution of that specific bandit. The total reward will be the weighted sum of these payoffs, $\sum_t w_t \times r_t$, where $r_t$ is the payoff received at time $t$ and $w_t$ is the associated weight. Your goal is to determine a strategy that will maximize this weighted sum. A standard model in this framework is the *infinite-horizon discounted-total-reward* model, where $w_t = \beta^t$ for some discount $\beta \in (0, 1)$. Under this model, there is an amazing result: At each time, let $P(c_i)$ represent the payoff distribution for bandit (coin) $c_i$; in general, this is conditioned on the outcomes of its previous results. We can compute a single real value $g_\beta(\, P(c_i)\,) \in \Re$ for each bandit, called its "Gittins index", and know that the optimal action is to pull the bandit $i^* = \arg\max_{i \in \mathcal{I}} g_\beta(\, P(c_i)\,)$ with the largest value [BF85, Git89]. Note that this $g_\beta(\, P(c_i)\,)$ depends only on the single bandit (*i.e.*, context independent[3]), and is able to incorporate both the "long term" rewards of learning more about this bandit, and the "immediate reward" of exploiting this bandit. This value $g_\beta(\, P(c_i)\,)$ corresponds to the constant-payoff of another "constant-valued" bandit, which is in a sense equivalent to $c_i$ (see [BF85]).

In our situation, with budget $b$, we may formate the problem in a discounted framework by setting $w_t = 0$ for $t = 1..b$, then $w_{b+1} = 1$ followed by $w_t = 0$ for $t > b + 1$. Hence, the first $b$ flips are pure exploration (as we do not receive any reward for these actions), so we can flip the best coin (using our observations) at time $b + 1$, Time $b + 1$ is pure exploitation. As in the budgeted problem we are not allowed further exploration, the discounts are set to 0 for time $t > b + 1$.

While our reward structure $\{w_t\}$ is significantly different from the infinite-horizon problem, we can use adapt the Gittins algorithm to our finite budget case (as in [SM02, Git89]), and observe how it performs. To do this, we set the value of the discount $\beta$ to compensate for the remaining budget: At each time, when the remaining budget is $s$, we set the discount $\beta_s$ so that the expected number of flips is $s$, that is $\beta_s = 1 - 1/s$. In our situation, therefore, we compute the Gittens index $g_{\beta_s}(\, \langle \alpha_1, \alpha_2 \rangle\,)$, for $\alpha_1$ and $\alpha_2$ (corresponding to the state of the coin $\Theta \sim B(\, \alpha_1,\, \alpha_2\,)$) and remaining budget $s$. We omit the details of computing these indices (see [SM02, MLG04]). At each time point, we choose the coin with the largest Gittins index for the number of flips $s$ remaining;

$$i^{GI} = \underset{i}{\operatorname{argmax}} \{\, g_{\beta_s}(\, \langle \beta_{i1}, \beta_{i2} \rangle\,)\,\}$$

[3]Unfortunately, no such context independent measure exists in our setting; see Example 2.

| Policy | Uses data? | Uses budget? |
|---|---|---|
| Round Robin | No | No |
| Random | No | No |
| Greedy | Yes | No |
| Biased Robin | Yes | No |
| SingleCoinLook | Yes | Yes |
| Interval Estim. | Yes | No |
| Gittins | Yes | Yes |

Table 1: Summary of Algorithms

#### 3.1.6 Summary

Table 1 summarizes some relevant properties of the algorithms.

## 4 Empirical Performance

We report on the performance of the algorithms in the important special case of identical costs and priors. We compute the optimal regret through exhaustive search for the range of only about $n \leq 10$ coins and budget $b \leq 10$. Figure 2(a) shows the performance of the optimal strategy against some of the other algorithms on uniform priors[4]. Note that on uniform priors, we can compare experimental regret averages on Round-Robin against the exact expectation given by Equation 1. The performances of look-ahead and Biased-Robin are very close to optimal. We have made similar observations on other types of identical priors on the same problem sizes (*e.g.*, Figure 2(b)). Figure 2(c)–(e) show the performances with the budget at $40$, $n = 10$. We computed the regrets at every intermediate time point to illustrate the performance of the algorithms as the budget is reached.

The Biased-Robin and look-ahead strategies consistently outperform the others, with look-ahead being the most time consuming algorithm. The relative difference in performance increases with priors skewed towards higher head probabilities (Beta densities $B(\, 5,\, 1\,)$ and $B(\, 10,\, 1\,)$ in the figure), and with increased $n$ (Figure 2(f); see also [MLG04]). The reason for the poor performance of Greedy is simple: due to its myopic property, it often cannot distinguish among different flips, and flips an arbitrary coin (the first coin in our implementation), and thus tends to regularly waste flips. For example, with uniform priors, as soon as some coin gives a head, a single flip of any coin does not change the expected highest mean. On the other hand, while look-ahead simplifies by considering single-coin allocations only, it performs well since it takes the whole budget into account.

Interval Estimation performs poorly especially in the case of skewed priors (Figure 2(d)). On such priors, Interval

[4]Each point is the average of at least 1000 trials. Initially in each trial, every coin's head probability is drawn from the prior, and then flipped by the algorithms as requested. Error bars are not shown for clarity (see [MLG04])

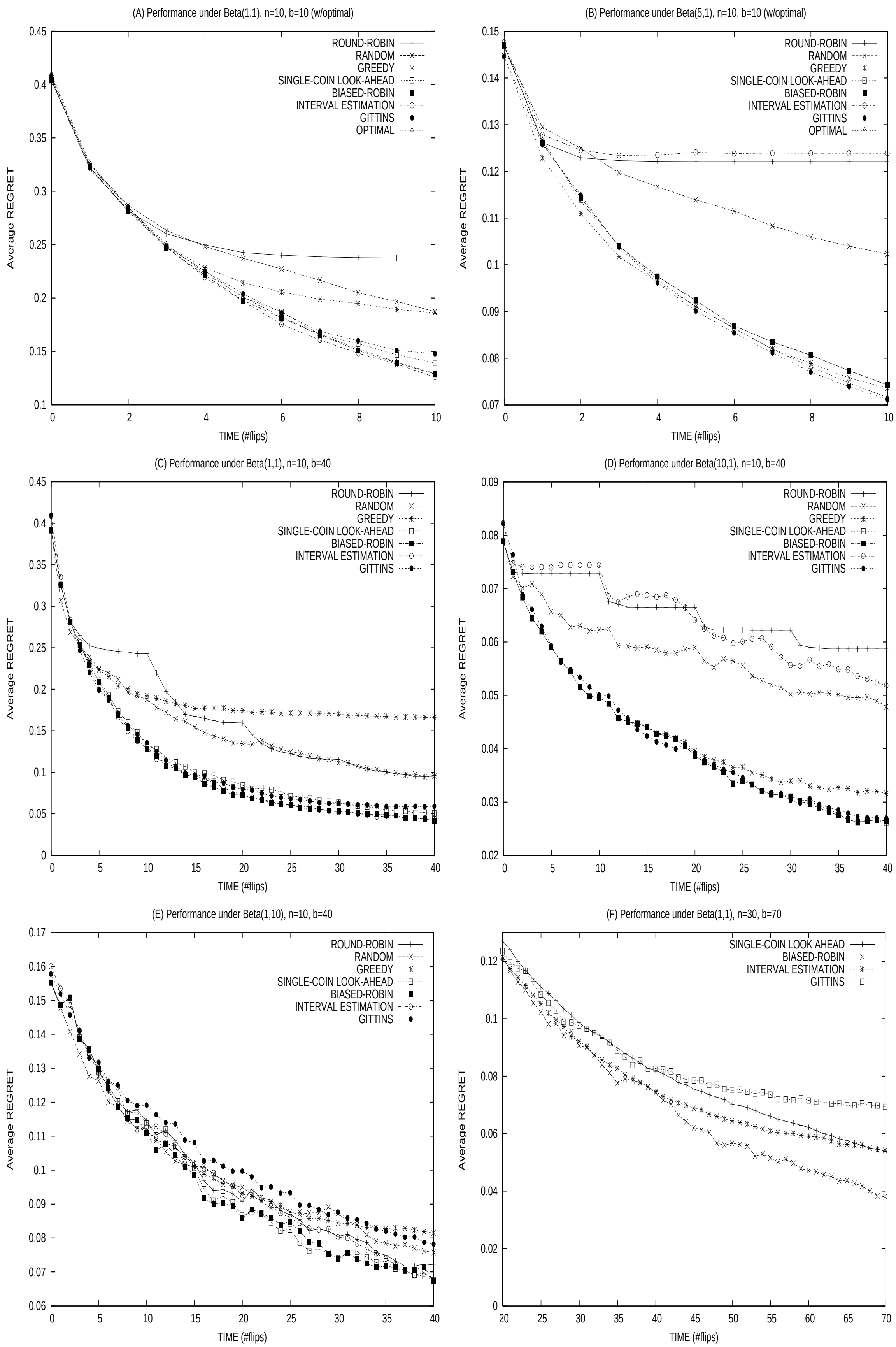

Figure 2: (a)–(e) The performance of the algorithms on 10 coins. (a)–(b) Compared against optimal, $0 \leq b \leq 10$, (a) Uniform priors, $B(1, 1)$. (b) Skewed $B(5, 1)$ (c) Uniform, $b = 40$ (d) Skewed $B(10, 1)$, $b = 40$ (e) Skewed $B(1, 10)$, $b = 40$ (f) Uniform, $n = 30, b = 70$

Estimation behaves like Round-Robin: when a coin yields a few consecutive heads, the algorithm moves to an untouched coin, as the upper end of its uncertainty region becomes smaller than that of an untouched coin! This observation suggests that explicitly accounting for uncertainty is not straightforward, and in particular setting the tolerance for uncertainty to a constant does not address the problem adequately. Furthermore, the observation that Interval Estimation can behave much like Round-Robin implies that it is not an approximation algorithm for the case of identical skewed priors.

While our version of Gittins indices algorithm (handling a finite budget) is designed for the total reward objective[5], we see that it performs reasonable. Still, it does not explore sufficiently, and it under-performs the algorithm we consider best: Biased-Robin. Single-coin look-ahead algorithm is also beaten (Figure 2(f)), for a similar reason: it often sticks to current best coin, but while a single untouched coin may not have a good chance of beating the current best, multiple such untouched coins may [MLG04]. Due to its performance, efficiency, and simplicity, the Biased-Robin algorithm is the algorithm of choice among the algorithms we tested. It is open whether it has an approximation guarantee.

## 5 Related work

There is a vast literature on sequential decision making, sample complexity of learning, active learning, and experiment design, all somewhat related to our work; we can only cite a few here. As noted in Section 3.1.5, our active model selection problem is an instance of the general class of the multi-armed bandit problem [BF85], which typically involves a trade-off between exploration (learning) and exploitation (reward accumulation). In our problem, there is a pure learning phase (determined by a fixed budget) followed by a pure exploitation instance; that is, there is no action rewards or costs, except at the final time point (the regret), where it is a function of the entire belief state (coin densities). In the typical bandit problem, there is an immediate reward obtained from action execution, which affects the objective just as the information gained does. These differences changes the nature of the task significantly, and sets it apart from typical finite or infinite-horizon bandit problems and their analyses (e.g., [KL00, EDMM02, HS02, ACBFS02]). To the best of our knowledge (and to our surprise) our budgeted problem has not been studied in the bandit literature in a computational framework before[6].

Our coins problem is also a special finite-horizon Markov decision problem (MDP) [Put94], but the state space in the direct formulation is too large to allow us to use standard MDP solution techniques. While the research on techniques for solving large MDPs with various forms of structures show promise (e.g., [Duf02]), we believe that our problem has special structure that allows for simpler, more efficient, and more effective algorithms for our special case.

This active model selection is very related to standard experimental design. Much of that work [CV95, BF85] involves a single allocation decision at the start of the testing phase (see Section 3.1.2): *e.g.*, 10 individuals receive treatment 1, 5 for treatment 2, and 25 for treatment 3. In that work, the learner (there, “experiment designer”) will commit to using that specific allocation — here for all 40 individuals. By contrast, our approach (including our allocation-bsed approaches) are used only to identify the *next single test* to perform; based on its outcome, the learner then decides what to do next, dynamically.

The coins problem is an instance of active learning and cost-sensitive learning (*e.g.*, [LMR02, Tur00, GGR02]). Feature costs in [Tur00, GGR02] refer to costs occuring at *classification* time, while we are concerned with costs during the learning phase. Similar to several results [LMR02, RM01], we show that *selective querying* can be much more efficient than a naïve method such as random. These previous results suggest that Greedy methods are effective; deeper look-aheads are not used due to a combination of inefficiency and non-significant gains [LMR02]. However, we observe in our setting that the Greedy method has poor performance both in theory and in experiments, while looking deeper pays significant dividends.

This paper considers an abstraction that allows us to obtain crisp theoretical results and many useful technical insights. Our related work [LMG03] takes a similar Bayesian approach and uses similar solution techniques to handle learning of Naïve Bayes classifiers under a budget. A significant difference here is in the formulation: here we are *selecting* from among a finite number of models, while there the objective is to *learn* the best model. However, that work shows that the basic algorithmic ideas presented here extend to yield effective selective querying algorithms in that context as well.

## 6 Future Work and Contributions

There are a number of directions for future work, in addition to the open problems already mentioned. An immediate extension of the problem is to select *non-trivial classifiers*. This problem may be formulated in a similar way: we are presented with $n$ candidate classifiers (imagine decision trees) with priors over a measure of their performance, and our task is to declare a winner. A significant additional dimension in this case is that querying a single feature af-fects not one but multiple classifiers in general. There are however "flatter" versions of the classifier problem, such as learning a Naïve Bayes classifier, in which these dependencies are limited. Our solution techniques more readily extend to the latter problems [LMG03].

[5] In fact, for $n = 30$, $b = 70$, and uniform priors, the average total accumulated reward is respectively 59, 58, 54, and 49 for Gittins, single-coin look-ahead, Biased-Robin, and Interval Estimation.

[6] Personal communication with D. Berry of [BF85].

**Contributions:** We introduced and motivated the active model selection problem. We investigated the computational complexity of the problem, and explored the performance of a variety of algorithms. Our analyses demonstrate significant problems with a number of algorithms. We presented a simple technique that significantly outperforms these alternatives.

## Acknowledgements

Many thanks to Jeff Schneider for suggesting the Gittins method and providing us the code. This research was supported by the Alberta Ingenuity Associateship, NSERC, Alberta Ingenuity Centre for Machine Learning, and Yahoo! Research Labs.